\documentclass[runningheads]{llncs}
\usepackage{graphicx}
\usepackage{amsmath,amssymb}
\usepackage{color}
\usepackage[width=122mm,left=12mm,paperwidth=146mm,height=193mm,top=12mm,paperheight=217mm]{geometry}
\usepackage{hyperref}

\usepackage{tikz}
\usepackage{booktabs}
\usepackage{comment}
\usepackage{subfig}
\usepackage{wrapfig}
\usepackage{xspace}
\usepackage{nicefrac}
\usepackage{float}
\usepackage[normalem]{ulem}

\makeatletter
\DeclareRobustCommand\onedot{\futurelet\@let@token\@onedot}
\def\@onedot{\ifx\@let@token.\else.\null\fi\xspace}
\def\eg{\emph{e.g}\onedot} 
\def\ie{\emph{i.e}\onedot}

\def\etal{\emph{et al}\onedot}
\makeatother

\newcommand{\est} [1] {$\widehat #1$}

\definecolor{col_wie_8}{RGB}{227,26,28}
\DeclareRobustCommand\mydot[1]{\tikz \node[circle,white, fill=col_wie_8, inner sep=0, font=\sffamily, circle,minimum size=0.2cm] {\footnotesize #1};}

\hypersetup{
  colorlinks   = true,
  linkcolor    = blue,
}

\newcommand{\new} [1] {#1}

\begin{document}
\pagestyle{headings}
\mainmatter

\title{Separating Reflection and Transmission Images in the Wild}

\titlerunning{Separating Reflection and Transmission Images in the Wild}

\authorrunning{Wieschollek, Gallo, Gu, and Kautz.}

\author{Patrick Wieschollek$^{1,2}$, Orazio Gallo$^1$, Jinwei Gu$^1$, and Jan Kautz$^1$}

\institute{$^1$NVIDIA, $^2$University of T\"{u}bingen}

\maketitle

\begin{abstract}

The reflections caused by common semi-reflectors, such as glass windows, can impact the performance of computer vision algorithms.
State-of-the-art methods can remove reflections on synthetic data and in controlled scenarios.
However, they are based on strong assumptions and do not generalize well to real-world images. Contrary to a common misconception, real-world images are challenging even when polarization information is used.
We present a deep learning approach to separate the reflected and the transmitted components of the recorded irradiance, which \emph{explicitly} uses the polarization properties of light.
To train it, we introduce an accurate synthetic data generation pipeline, which simulates realistic reflections, including those generated by curved and non-ideal surfaces, non-static scenes, and high-dynamic-range scenes.
 \end{abstract}

\begin{center}
    \centering
    \begin{tikzpicture}
      \node {\includegraphics[width=\textwidth]{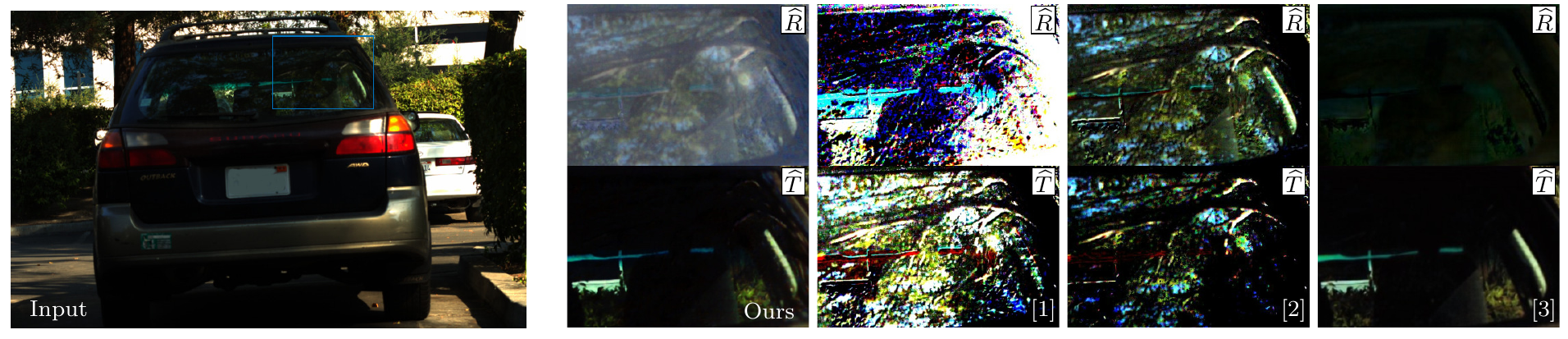}};

    \end{tikzpicture}
    \small{\captionof{figure}{Glass surfaces are virtually unavoidable in real-world pictures.
    Our approach to separate the reflection and transmission layers, works even for general, curved surfaces, which break the assumptions of state-of-the-art methods. In this example, only our method can correctly estimate both reflection $\widehat{R}$ (the tree branches) and transmission $\widehat{T}$ (the car's interior).}\label{fig:teaser}
    }
\end{center}

\section{Introduction}\label{sec:intro}
\nocite{projectpage}

\begin{figure}
\centering
 \resizebox{\linewidth}{!}{
\begin{tikzpicture}
	\node{\includegraphics[width=\textwidth]{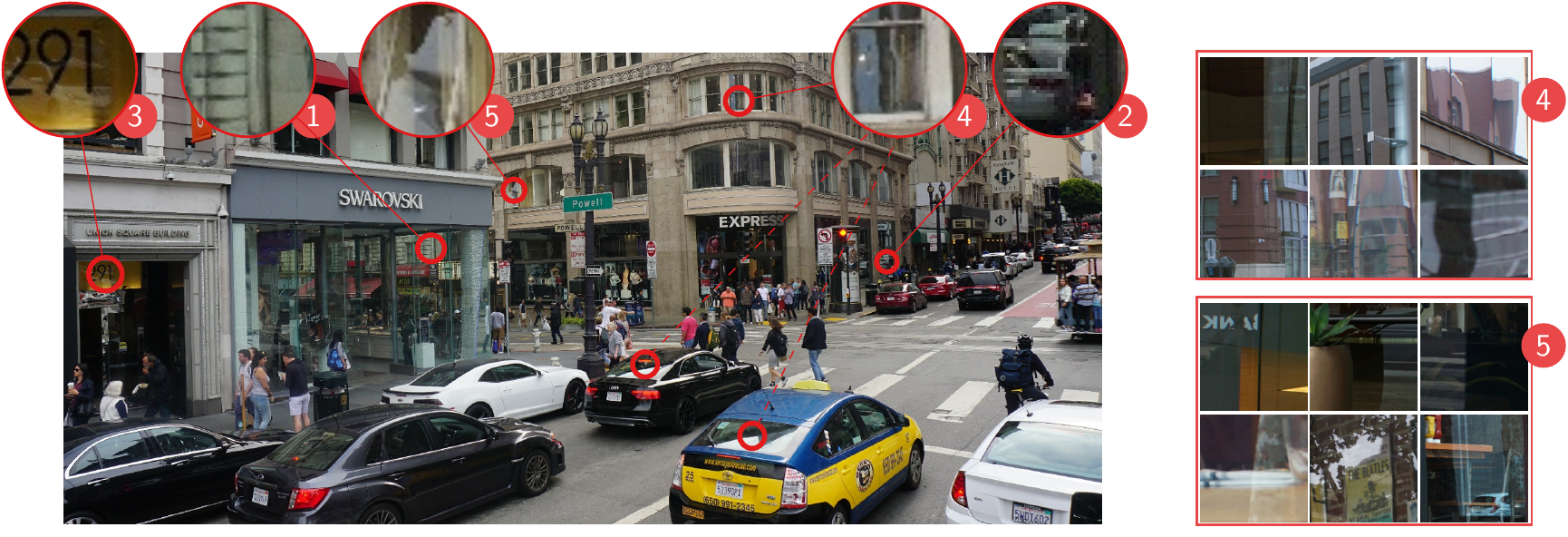}};
	\node at (-6.1, -1.8) {(a)};
	\node at (2.8, -1.8) {(b)};

\end{tikzpicture}
}
\small{\caption{
Depending on the ratio between transmitted and reflected radiance, a semi-reflector may produce no reflections \mydot{1}, pure reflections \mydot{2}, or a mix of the two, which can vary smoothly \mydot{3}, or abruptly \mydot{5}. The local curvature of the surface can also affect the appearance of the reflection \mydot{4}. The last two, \mydot{4} and \mydot{5}, are all but uncommon, as shown in (b).}\label{fig:case_study}}

\end{figure}

Computer vision algorithms generally rely on the assumption that the value of each pixel is a function of the radiance of a single area in the scene. Semi-reflectors, such as typical windows or glass doors, break this assumption by creating a superposition of the radiance of two different objects: the one behind the surface and the one that is reflected.
It is virtually impossible to avoid semi-reflectors in man-made environments, as can be seen in Figure~\ref{fig:case_study}(a), which shows a typical downtown area.
Any multi-view stereo or SLAM algorithm would be hard-pressed to produce accurate reconstructions on this type of images.

Several methods exist that attempt to separate the reflection and transmission layers.

\new{At a semi-reflective surface, the observed image can be modeled as a linear combination of the reflection and the transmission images: $I_o = \alpha_r I_r + \alpha_t I_t$.
The inverse problem is ill-posed as it requires estimating multiple unknowns from a single observation.}
A solution, therefore, requires additional priors or data. Indeed, previous works rely on assumptions about the appearance of the reflection (\eg, it is blurry), about the shape and orientation of the surface (\eg, it is perfectly flat and exactly perpendicular to the principal axis of the camera), and others.
Images taken in the wild, however, regularly break even the most basic of these assumptions, see Figure~\ref{fig:case_study}(b), causing the results of state-of-the-art methods~\cite{schechner2000polarization,kong2014pami,fan2017iccv} to deteriorate even on seemingly simple cases, as shown in Figure~\ref{fig:teaser}, which depicts a fairly typical real-world scene.

\new{One particularly powerful tool is is polarization: images captured through a polarizer oriented at different angles offer additional observations.}
Perhaps surprisingly, however, our analysis of the state-of-the-art methods indicates that the quality of the results degrades significantly when moving from synthetic to real data, \emph{even when using polarization}.
This is due to the simplifying assumptions that are commonly made, but also to an inherent issue that is all too often neglected: a polarizer's ability to attenuate reflections greatly depends on the viewing angle~\cite{collett2005field}.
The attenuation is maximal at an angle called the Brewster angle, $\theta_{B}$. However, even when part of a semi-reflector is imaged at $\theta_{B}$, the angle of incidence in other areas is sufficiently different from $\theta_{B}$ to essentially void the effect of the polarizer, as clearly shown in Figure~\ref{fig:notsoeasy}.
Put differently, because of the limited signal-to-noise ratio, for certain regions in the scene, \emph{the additional observations may not be independent}.

We present a deep-learning method capable of separating the reflection and transmission components of images captured \emph{in the wild}.
The success of the method stems from our two main contributions. First, rather than requiring a network to learn the reflected and transmitted images directly from the observations, we leverage the properties of light polarization and use a residual representation, in which the input images are projected onto the canonical polarization angles (Section~\ref{subsec:method_polarization} and~\ref{subsec:method_recovering}).
Second, we design an image-based data generator that faithfully reproduces the image formation model (Section~\ref{subsec:method_data}).

We show that our method can successfully separate the reflection and transmission layers even in challenging cases, on which previous works fail. To further validate our findings, we capture the Urban Reflections Dataset, a polarization-based dataset of reflections in urban environments that can be used to test reflection removal algorithms on realistic images.
Moreover, to perform a thorough evaluation against state-of-the-art methods whose implementation is not publicly available, we re-implemented several representative methods.
As part of our contribution, we release those implementations for others to be able to compare against their own methods~\cite{projectpage}.

\section{Related Work}\label{sec:related}

There is a rich literature of methods dealing with semi-reflective surfaces, which can be organized in three main categories based on the assumptions they make.

\emph{Single-image methods} can leverage gradient information to solve the problem. Levin and Weiss, for instance, require manual input to separate gradients of the reflection and the transmission~\cite{levin2007}.
Methods that are fully automated can distinguish the gradients of the reflected and transmitted images by leveraging the defocus blur~\cite{li2014single}: reflections can be blurry because the subject behind the semi-reflector is much closer than the reflected image~\cite{fan2017iccv}, or because the camera is focused at infinity and the reflected objects are close to the surface~\cite{arvanitopoulos2017single}.
Moreover, for the case of double-pane or thick windows, the reflection can appear ``doubled''~\cite{diamant2008overcoming}, and this can be used to separate it from the transmitted image~\cite{shih2015cvpr}.
While these methods show impressive results, their assumptions are stringent and do not generalize well to real-world cases, causing them to fail on common cases.

\emph{Multiple images captured from different viewpoints} can also be used to remove reflections.
Several methods propose different ways to estimate the relative motion of the reflected and transmitted image, which can be used to separate them~\cite{li2013iccv,xue2015siggraph,szeliski2000layer,guo2014cvpr,han2017cvpr}.
It is important to note that these methods assume static scenes---the motion is the apparent motion of the reflected layer relative to the transmitted layer, not scene motion.
Other than that, these methods make assumptions that are less stringent than those made by single-image methods.
Nonetheless, these algorithms work well when reflected and transmitted scenes are shallow in terms of depth, so that their velocity can be assumed uniform.
For the case of spatially and temporally varying mixes, Kaftory and Zeevi propose to use sparse component analysis instead~\cite{kaftory2013blind}.

\emph{Multiple images captured under different polarization angles} offer a third venue to tackle this problem.
Assuming that images taken at different polarization angles offer independent measurements of the same scene, reflection and transmission can be separated using independent component analysis~\cite{farid1999cvpr,barros2001ieice,bronstein2005sparse}.
\new{An additional prior that can be leveraged is given by double reflections, when the semi-reflective surface generates them~\cite{diamant2008overcoming}.}
Under ideal conditions, and leveraging polarization information, a solution can also be found in closed form~\cite{schechner2000polarization,kong2014pami}.
In our experiments, we found that most of the pictures captured in unconstrained settings break even the well-founded assumptions used by these papers, as shown in Figure~\ref{fig:case_study}.

\section{Method}\label{sec:method}

We address the problem of layer decomposition by leveraging the ability of a semi-reflector to polarize the reflected and transmitted layers differently.
Capturing multiple polarization images of the same scene, then, offers partially independent observations of the two layers.
\new{To use this information, we take a deep learning approach.}
Since the ground truth for this problem is virtually impossible to capture, we synthesize it.
As for any data-driven approach, the realism of the training data is paramount to the quality of the results.
In this section, after reviewing the image formation model, we give an overview of our approach, we discuss the limitations of the assumptions that are commonly made, and how we address them in our data generation pipeline. Finally, we describe the details of our implementation.
\subsection{Polarization, Reflections, and Transmissions}\label{subsec:method_polarization}
\begin{figure}
  \centering
  \begin{tikzpicture}
    \node{\includegraphics[width=.6\columnwidth]{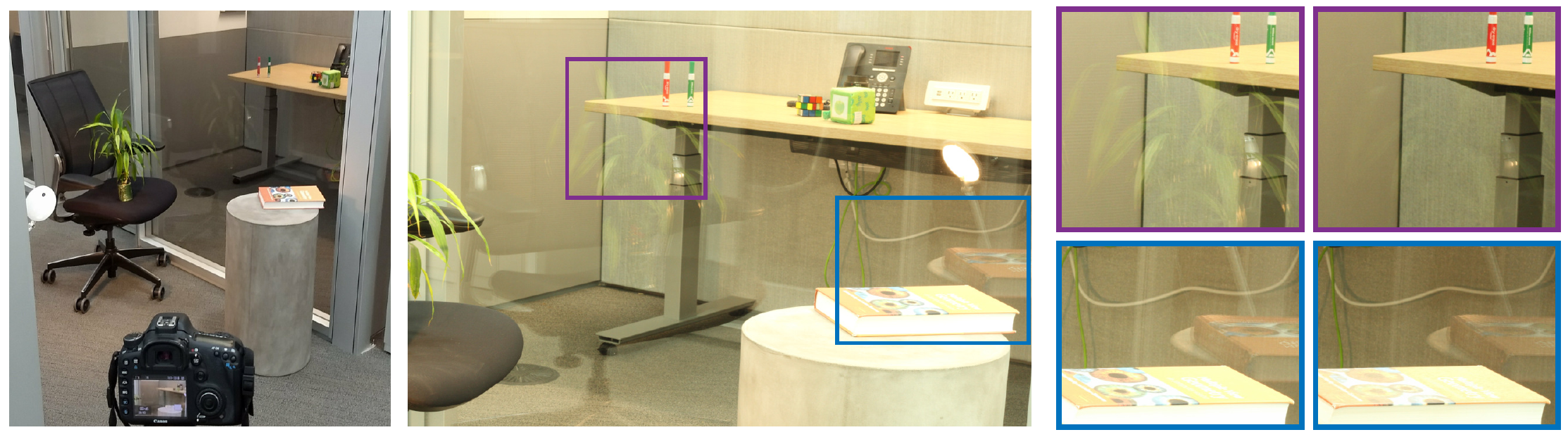}};
    \node at (1.9, -1.2) {$\phi\approx\phi_\perp$};
    \node at (3.1, -1.2) {$\phi\approx\phi_\Vert$};
  \end{tikzpicture}
  \includegraphics[width=.37\columnwidth]{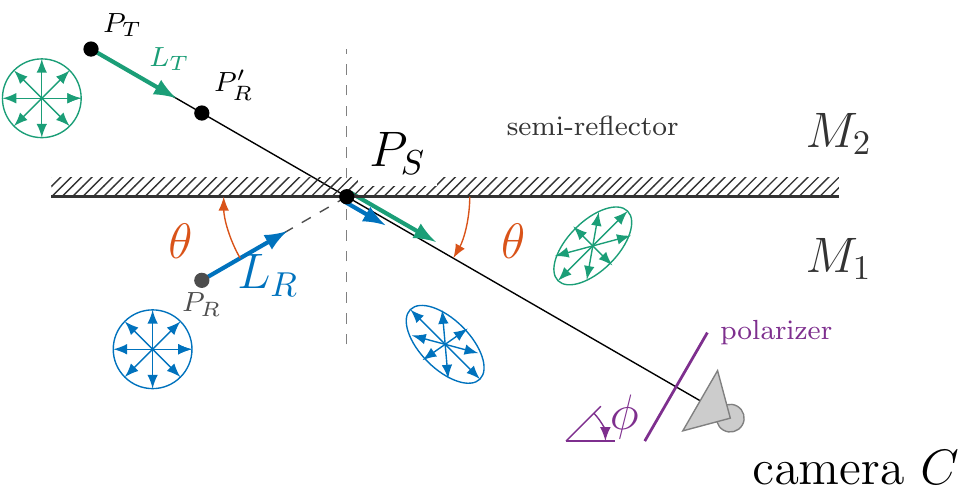}
    \caption{A polarizer attenuates reflections when they are viewed at the Brewster angle {$\theta=\theta_B$}. For the scene shown on the left, we manually selected the two polarization directions that maximize and minimize reflections respectively. Indeed, the reflection of the plant is almost completely removed. However, only a few degrees away from the Brewster angle, the polarizer
    has little to no effect, as is the case for the reflection of the book on the right.}
  \label{fig:notsoeasy}
\end{figure}

Consider two points, $P_R$ and $P_T$ such that $P^{'}_R$, the reflection of $P_R$, lies on the line of sight of $P_T$, and assume that both emit unpolarized light, {see Figure~\ref{fig:notsoeasy}}.
After being reflected or transmitted, unpolarized light becomes polarized by an amount that depends on $\theta$, the \emph{angle of incidence} (AOI).

\new{At point $P_S$, the intersection of the line of sight and the surface, the total radiance $L$ is a combination of the reflected radiance $L_R$, and the transmitted radiance $L_T$. Assume we place a linear polarizer with polarization angle $\phi$ in front of the camera. If we integrate over the exposure time, the intensity at \textit{each pixel} $x$ is}
\begin{align}
  I_{\phi}(x) = \alpha(\theta, \phi_\perp, \phi) \cdot \frac{I_R(x)}{2} + \left(1-\alpha(\theta, \phi_\perp, \phi)\right) \cdot \frac{I_T(x)}{2},
  {}\label{eq:polarization_model}
\end{align}
where the mixing coefficient $\alpha(\cdot) \in [0,1]$, the angle of incidence $\theta(x)  \in [0, \nicefrac{\pi}{2}]$, the $p-$polarization direction~\cite{schechner2000polarization} $\phi_{\perp}(x)\in [-\nicefrac{\pi}{4}, \nicefrac{\pi}{4}]$, and the reflected and transmitted images \emph{at the semi-reflector}, $I_R(x)$ and $I_T(x)$, are all unknown.

At the Brewster angle, $\theta_{B}$, the reflected light is completely polarized along $\phi_{\perp}$, \ie in the direction perpendicular to the incidence plane\footnote{The incidence plane is defined by the direction in which the light is traveling and the semi-reflector's normal.}, and the transmitted light along $\phi_{\parallel}$, the direction parallel to the plane of incidence. The angles $\phi_{\perp}$ and $\phi_{\parallel}$ are called the \emph{canonical} polarization angles.
In the unique condition in which $\theta(x) = \theta_{B}$, two images captured with the polarizer at the canonical polarization angles offer independent observations that are sufficient to disambiguate between $I_R$ and $I_T$. Unless the camera or the semi-reflector are at infinity, however, $\theta(x) = \theta_{B}$ only holds for few points in the scene, if any, as shown in Figure~\ref{fig:notsoeasy}.
To complicate things, for curved surfaces, $\theta(x)$ varies non-linearly with $x$.
Finally, even for arbitrarily many acquisitions at different polarization angles, $\phi_{j}$, the problem remains ill-posed as each observation $I_{\phi_{j}}$ adds new pixel-wise unknowns $\alpha(\theta,\phi_{\perp},\phi_j)$.

\subsection{Recovering $R$ and $T$}
\label{subsec:method_recovering}

\begin{comment}

\end{comment}

\begin{figure}[tb]
  \centering
  \includegraphics[width=0.85\textwidth]{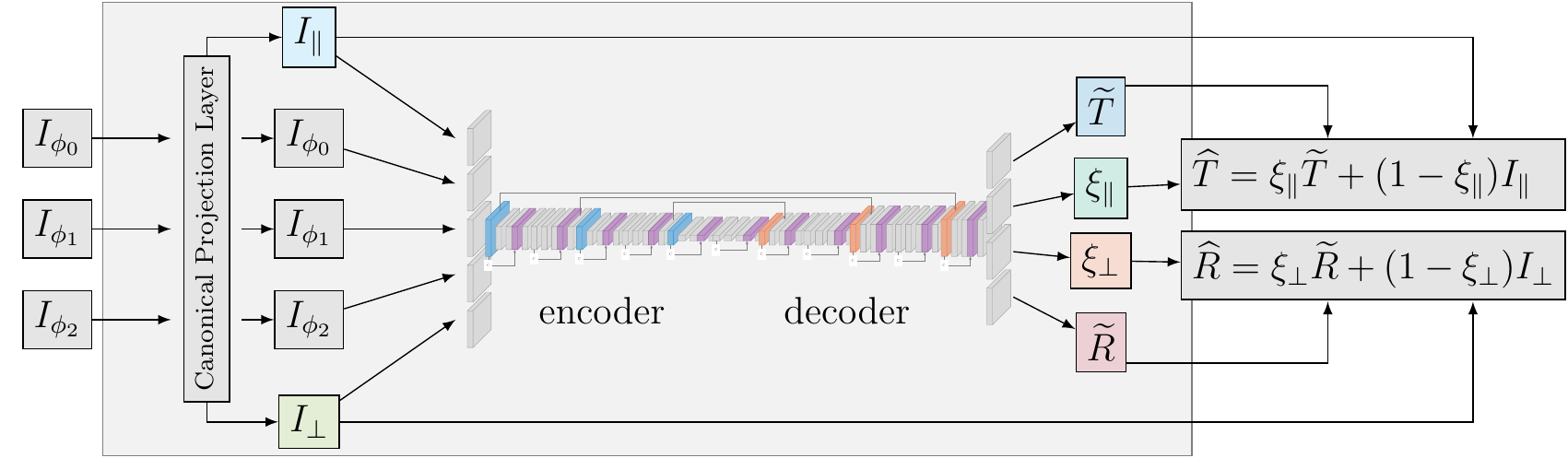}
  \caption{Our encoder-decoder network architecture with ResNet blocks includes a Canonical Projection Layer, which projects the input images onto the canonical polarization directions, and uses a residual parametrization for $\widehat{T}$ and $\widehat{R}$.}
  \label{fig:networkdesign}
  \vspace{-0.7cm}
\end{figure}

When viewed through a polarizer oriented along direction $\phi$, $I_R$ and $I_T$, which are the reflected and transmitted images \emph{at the semi-reflector}, produce image $I_\phi$ \emph{at the sensor}.
Due to differences in dynamic range, as well as noise, in some regions the reflection may dominate $I_\phi$, or vice versa, see Section~\ref{sec:dynamic_range}.
Without hallucinating content, one can only aim at separating $R$ and $T$, which we define to be the observable reflected and transmitted components.
For instance, $T$ may be zero in regions where $R$ dominates, even though $I_T$ may be greater than zero in those regions.
To differentiate them from the ground truth, we refer to our estimates as \est{R} and \est{T}.

To recover \est{R} and \est{T}, we use an encoder-decoder architecture, which has been shown to be particularly effective for a number of tasks, such as image-to-image translation~\cite{pix2pix2017}, denoising~\cite{NIPS2016Mao}, or deblurring~\cite{iccv2017/Wieschollek}.
Learning to estimate \est{R} and \est{T} directly from images taken at arbitrary polarization angles does not produce satisfactory results.
One main reason is that parts of the image may be pure reflections, thus yielding no information about the transmission, and vice versa.

To address this issue, we turn to the polarization properties of reflected and transmitted images.
Recall that $R$ and $T$ are maximally attenuated, though generally not completely removed, at $\phi_{\Vert}$ and $\phi_{\perp}$ respectively.
The canonical polarization angles depend on the geometry of the scene, and are thus hard to capture directly.
However, we note that an image $I_{\phi}(x)$ can be expressed as~\cite{kong2014pami}:
\begin{equation}\label{eq:projecting}
I_{\phi}(x) = I_{\perp}(x)\cos^2(\phi - \phi_{\perp}(x)) + I_{\Vert}(x) \sin^2(\phi - \phi_{\perp}(x)).
\end{equation}
Since Equation~\ref{eq:projecting} has three unknowns, $I_{\perp}$, $\phi_{\perp}$, and $I_{\Vert}$, we can use three different observations of the same scene, $\left\{I_{\phi_i}(x)\right\}_{i=\{0,1,2\}}$, to obtain a linear system that allows to compute $I_{\perp}(x)$ and $I_{\Vert}(x)$. To further simplify the math we capture images such that $\phi_i = \phi_0+i\cdot \nicefrac{\pi}{4}$.

For efficiency, we implement the projection onto the canonical views as a network layer in TensorFlow. The canonical views and the actual observations are then stacked in a 15-channel tensor and used as input to our network.
Then, instead of training the network to learn to predict \est{R} and \est{T}, we train it to learn the \emph{residual} reflection and transmission layers.
More specifically, we train the network to learn an 8-channel output, which comprises the residual images $\widetilde{T}(x)$, $\widetilde{R}(x)$, and the two single-channel weights $\xi_\Vert(x)$ and $\xi_\perp(x)$. Dropping the dependency on pixel $x$ for clarity, we can then compute:
\begin{eqnarray}
\widehat{R} = \xi_\perp \widetilde{R} + (1-\xi_\perp)I_\perp\qquad \text{and}\qquad
\widehat{T} = \xi_\Vert \widetilde{T} + (1-\xi_\Vert)I_\Vert.
\end{eqnarray}
While $\xi_\perp$ and $\xi_\Vert$ introduce two additional unknowns per pixel, they significantly simplify the prediction task in regions where the canonical projections are already good predictors of \est{R} and \est{T}.
We use an encoder-decoder with skip connections~\cite{ronneberger2015unet} that consists of three down-sampling stages, each with two ResNet blocks~\cite{he2016resnet}. The corresponding decoder mirrors the encoding layers using a transposed convolution with two ResNet blocks.
We use an $\ell_2$ loss on \est{R} and \est{T}. We also tested $\ell_1$ and a combination of $\ell_1$ and $\ell_2$, which did not yield significant improvements.

The use of the canonical projection layer, as well as the parametrization of residual images is key for the success of our method. We show this in the {Supplementary}, where we compare the output of our network with the output of the exact same architecture trained to predict \est{R} and \est{T} directly from the three polarization images $I_{\phi_i}(x)$.

\begin{comment}

\end{comment}
 \subsection{Image-Based Data Generation}\label{subsec:method_data}
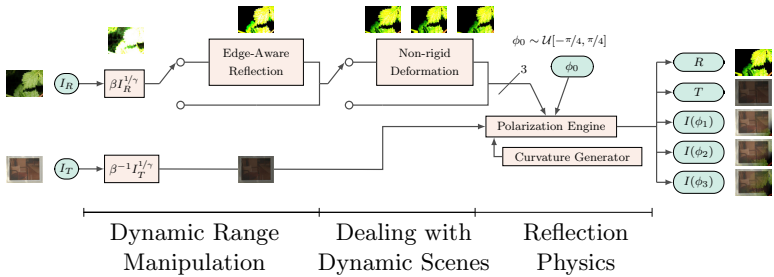
\begin{figure}[t]
  \centering
  \begin{tikzpicture}[
  annotation/.style={rectangle, draw=red!40, inner sep=1, fill=red!20}]
    \node {\includegraphics[width=0.85\textwidth]{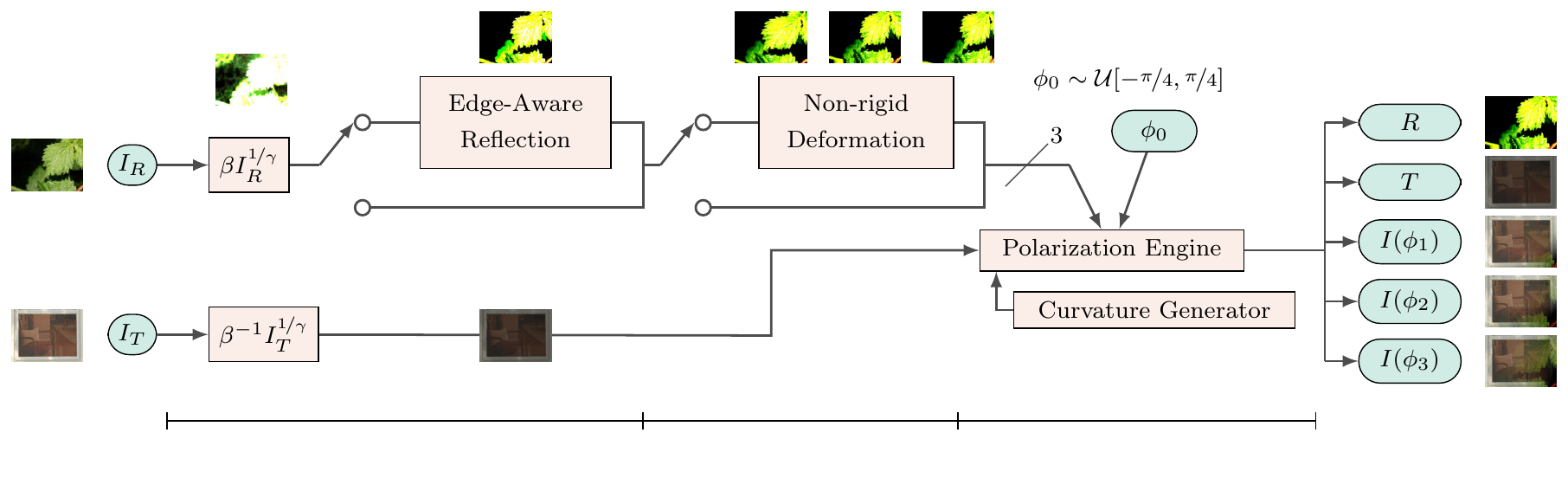}};

    \node[align=center] at (-2.6,-1.6) {\footnotesize Dynamic Range \\Manipulation};
    \node[align=center] at (0.2,-1.6) {\footnotesize Dealing with\\ Dynamic Scenes};
    \node[align=center] at (2.5,-1.6) {\footnotesize Reflection\\Physics};
  \end{tikzpicture}
  \caption{Our image-based data generation procedure.
  We apply several steps to images $I_R$ and $I_T$ simulating reflections in most real-world scenarios (Section~\ref{sec:dynamic}).
  }
  \label{fig:flow_chart_data}
  \vspace{-0.7cm}
\end{figure}

The ground truth data to estimate \est{R} and \est{T} is virtually impossible to capture in the wild.
Recently, Wan~\etal released a dataset for single-image reflection removal~\cite{wan2017benchmarking}, but it does not offer polarization information.
In principle, Equation~\ref{eq:polarization_model} could be used directly to generate, from any two images, the data we need. The term $\alpha$ in the equation, however, hides several subtleties and nonidealities. For instance, previous polarization-based works use it to synthesize data by assuming uniform AOI, perfectly flat surfaces, comparable power for the reflected and transmitted irradiance, or others. This generally translates to poor results on images captured in the wild: Figures~\ref{fig:teaser} and~\ref{fig:case_study} show common scenes that violate all of these assumptions.

We propose a more accurate synthetic data generation pipeline, see Figure~\ref{fig:flow_chart_data}. Our pipeline starts from two randomly picked images from the PLACE2 dataset~\cite{zhou2017places}, $I_R$ and $I_T$, which we treat as the image of reflected and transmitted scene \emph{at the surface}. From those, we model the behaviors observed in real-world data, which we describe as we ``follow'' the path of the photons from the scene to the camera.
\subsubsection{Dynamic Range Manipulation at the Surface}\label{sec:dynamic_range}

To simulate realistic reflections, the dynamic range (DR) of the transmitted and reflected images \emph{at the surface} must be significantly different. This is because real-world scenes are generally high-dynamic-range (HDR). Additionally, the light intensity at the surface drops with the distance from the emitting object, further expanding the combined DR.
However, our inputs are low-dynamic-range images because a large dataset of HDR images is not available.
We propose to artificially manipulate the DR of the inputs so as to match the appearance of the reflections we observe in real-world scenes.

Going back to Figure~\ref{fig:notsoeasy} (right), we note that for regions where
 $L_T \approx L_R$, a picture taken without a polarizer will capture a smoothly varying superposition of the images of $P_R$ and $P_T$ (Figure~\ref{fig:case_study} \mydot{3}). For areas of the surface where $L_R \gg L_T$, however, the total radiance is $L \approx L_R$, and the semi-reflector essentially acts as a mirror (Figure~\ref{fig:case_study} \mydot{2}). The opposite situation is also common (Figure~\ref{fig:case_study} \mydot{1}).
To allow for these distinct behaviors, we manipulate the dynamic range of the input images with a random factor $\beta\sim \mathcal{U}[1, K]$:
\begin{eqnarray}
\tilde{I}_R =  \beta I_R^{1/\gamma}\qquad \text{and} \qquad \tilde{I}_T =  \frac{1}{\beta} I_T^{1/\gamma},\label{eq:two}
\end{eqnarray}
where $1/\gamma$ linearizes the gamma-compressed inputs\footnote{Approximating the camera response function with a gamma function does not affect the accuracy of our results, as we are not trying to produce data that is radiometrically accurate with respect to the original scenes.}. We impose that $K > 1$ to compensate for the fact that a typical glass surface transmits a much larger portion of the incident light than it reflects\footnote{At an angle of incidence of $\nicefrac{\pi}{4}$, for instance, a glass surface reflects less than $16\%$ of the incident light.}.

Images $\tilde{I}_R$ and $\tilde{I}_T$ can reproduce the types of reflections described above, but are limited to those cases for which $L_R - L_T$ changes smoothly with $P_S$.
However, as shown in
Figure~\ref{fig:case_study} \mydot{5},
the reflection can drop abruptly following the boundaries of an object. This may happen when an object is much closer than the rest of the scene, or when its radiance is larger than the surrounding objects.
To properly model this behavior, we treat it as a type of reflection on its own, which we apply to a random subset of the image whose range we have already expanded.
Specifically, we set to zero the regions of the reflection or transmission layer, whose intensity is below $T =  \text{mean}(\tilde{I}_R+\tilde{I}_T)$, similarly to the method proposed by Fan~\etal~\cite{fan2017iccv}.

\subsubsection{Dealing with Dynamic Scenes}\label{sec:dynamic}
Our approach requires images captured under three different polarization angles. While cameras that can simultaneously capture multiple polarization images exist~\cite{fluxdata,ricoh,polarcam}, they are not widespread. To date, the standard way to capture different polarization images is sequential; this causes complications for non-static scenes.
As mentioned in Section~\ref{sec:related}, if multiple pictures are captured from different locations, the relative motion between the transmitted and reflected layers can help disambiguate them.
Here, however, ``non-static'' refers to the scene itself, such as is the case when a tree branch moves between the shots.
Several approaches were proposed that can deal with dynamic scenes in the context of stack-based photography~\cite{gallo2016stack}.
Rather than requiring some pre-processing to fix artifacts due to small scene changes at inference time, however, we propose to synthesize training data that simulates them, such as local, non-rigid deformations.
We first define a regular
 grid over a patch, and then we perturb each one of the grid's anchors by $(dx, dy)$, both sampled from a Gaussian with variance $\sigma_{\text{NR}}^2$
 , which is also drawn randomly for each patch.
 We then interpolate the position of the rest of the pixels in the patch.
For each input patch, we generate three different images, one per polarization angle. We only apply this processing to a subset of the synthesized images---the scene is not always dynamic. Figure~\ref{fig:synth_data}(a) and (b) show an example of original and distorted patch respectively.

\subsubsection{Geometry of the Semi-Reflective Surface}\label{sec:surface}
The images synthesized up to this point can be thought of as the irradiance of the unpolarized light at the semi-reflector. After bouncing off of, or going through, the surface, light becomes polarized as described in Section~\ref{subsec:method_polarization}. The effect of a linear polarizer placed in front of the camera and oriented at a given polarization angle, depends on the angle of incidence (AOI) of the \emph{specific} light ray. Some previous works assume this angle to be uniform over the image, which is only true if the camera is at infinity, or if the surface is flat.

We observe that real-world surfaces are hardly ever perfectly flat. Many common glass surfaces are in fact designed to be curved, as is the case of car windows, see Figure~\ref{fig:teaser}. Even when the surfaces are meant to be flat, the imperfections of the glass manufacturing process introduce local curvatures, see
Figure~\ref{fig:case_study}~\mydot{4}.

\begin{figure}[t]
  \centering
  \begin{tikzpicture}
    \node at (1, -2.0) {\includegraphics[width=.12\columnwidth]{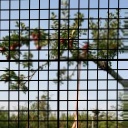}};
    \node at (1, -4.2) {\includegraphics[width=.12\columnwidth]{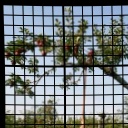}};
    \node at (4.2, -3) {\includegraphics[width=.35\textwidth]{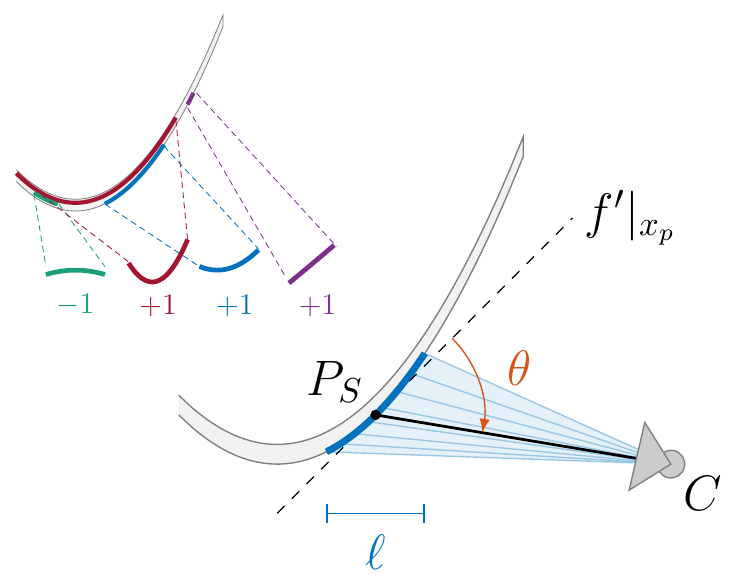}};
    \node at (9.5, -3) {\includegraphics[width=.49\textwidth]{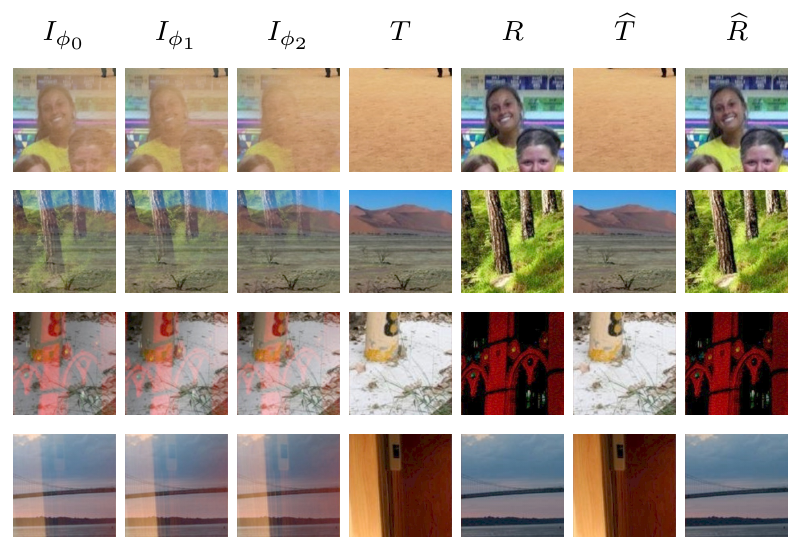}};
    \node at (1, -3) {(a)};
    \node at (1, -5.2) {(b)};
    \node at (4.2, -5.2) {(c)};
    \node at (9.5, -5.2) {(d)};
  \end{tikzpicture}

  \caption{Examples of our non-rigid motion deformation (a, b) and a curved surface-generator given the camera position, $C$, a surface-point, $P_S$, length, $\ell$, and the convexity $\pm1$ (c).
  Randomly sampled training data (d) with synthesized observations $I_{{\phi}_0}$, $I_{{\phi}_1}$, $I_{{\phi}_2}$ from the ground truth data $T$ and $R$, and estimates $\widehat{T}, \widehat{R}$.
  }
  \label{fig:synth_data}
\end{figure}
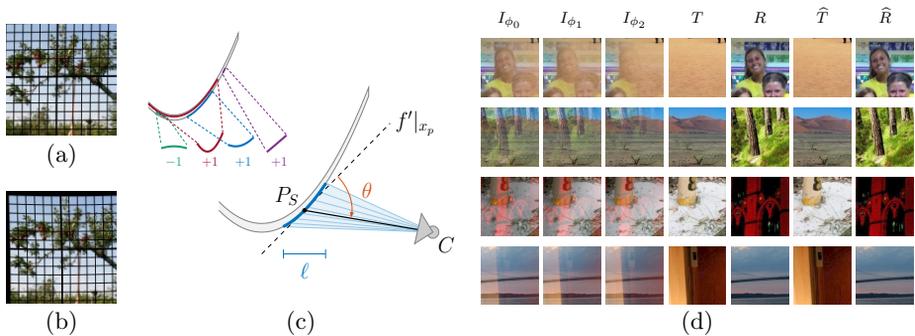

At training time, we could generate unconstrained surface curvatures to account for this observation. However, it would be difficult to sample realistic surfaces. Moreover, the computation of the AOI from the surface curvature may be non-trivial.
As a regularizer, we propose to use a parabola. When the patches are synthesized, we just sample four parameters: the camera position $C$, a point on the surface $P_S$, a segment length, $\ell$, and the convexity as $\pm 1$, Figure~\ref{fig:synth_data}(c). Since the segment is always mapped to the same output size, this parametrization allows to generate a number of different, realistic curvatures.
Additionally, because we use a parabola, we can quickly compute the AOI in closed form, from the sample parameters, see Supplementary.

\subsection{Implementation Details}\label{subsec:method_network}
From the output of the pipeline described so far, the simulated AOI, and a random polarization angle $\phi_0$, the polarization engine generates three observations with polarization angles separated by $\nicefrac{\pi}{4}$, see Figure~\ref{fig:flow_chart_data}.
In practice, the polarizer angles $\phi_i$ will be inaccurate for real data due to the manual adjustments of the polarizer rotation. We account for this by adding noise within $\pm 4^\circ$ to each polarizer angle $\phi_i$.
Additionally we set $\beta \sim \mathcal{U}[1, 2.8]$.
The input for our neural network is $\mathbb{R}^{B\times 128\times 128\times 9}$ when trained on $128\times 128$ patches, where $B=32$ is the batch size. We trained the model from scratch with a learning rate $5\cdot 10^{-3}$ using ADAM. See the Supplementary for more details about the architecture.
The colors of the network predictions might be slightly desaturated~\cite{colorissue2016wieschollek,superres2016kim,fan2017iccv}. We use a parameter-free color-histogram matching against one of the observations to obtain the final results.

\section{Experiments}\label{sec:results}

In this section we evaluate our method and data modeling pipeline on both synthetic and real data.
For the latter, we introduce the Urban Reflections Dataset (URD), a new dataset of images containing semi-reflectors captured with polarization information.
\new{A fair evaluation can only be done against other polarization-based methods, which use multiple images.
However, we also compare against single-image methods for completeness.}

\paragraph{The Urban Reflections Dataset (URD).} For practical relevance, we compile a dataset of 28 high-resolution RAW images (24MP) that are taken in urban environments using two different consumer cameras (Alpha 6000 and Canon EOS 7D, both ASP-C sensors), and which we make publicly available. The Supplementary shows all the pictures in the dataset.
This dataset includes examples taken with a wide aperture, and while focusing on the plane of the semi-reflector, thus meeting the assumptions of Fan~\etal\cite{fan2017iccv}.

\subsection{Numerical Performance Evaluation}

\begin{wraptable}{r}{0.5\textwidth}
  \caption{Cross-validation on synthetic data. Best results in bold.}
  \label{tab:synthetic_eval}
  \centering
\resizebox{\linewidth}{!}{
  \begin{tabular}{lp{1.5cm}p{1.5cm}cp{1.5cm}p{1.5cm}}
  \toprule
                                                  & \multicolumn{2}{c}{PASCAL VOC 2012} && \multicolumn{2}{c}{PLACE2} \\
                                                   \cline{2-3}  \cline{5-6}
  Method                                          & RMSE                                & PSNR                          && RMSE  & PSNR\\
  \midrule
  Farid~\etal\cite{farid1999cvpr}                 & 0.401                               & 7.93                          && 0.380 & 8.38\\
  Kong~\etal\cite{kong2014pami}                   & 0.160                               & 15.88                         && 0.156 & 16.12\\
  Schechner~\etal\cite{schechner2000polarization} & 0.085                               & 21.34                         && 0.086 & 21.27\\
  Fan~\etal\cite{fan2017iccv}                     & 0.080                               & 21.89                         && 0.084 & 21.48\\
  Ours                                            & \textbf{0.064}                               & \textbf{23.83}                         && \textbf{0.066} & \textbf{23.58}\\
  \bottomrule
  \end{tabular}
}
\end{wraptable}

\begin{figure}
\centering
\includegraphics[width=1\textwidth]{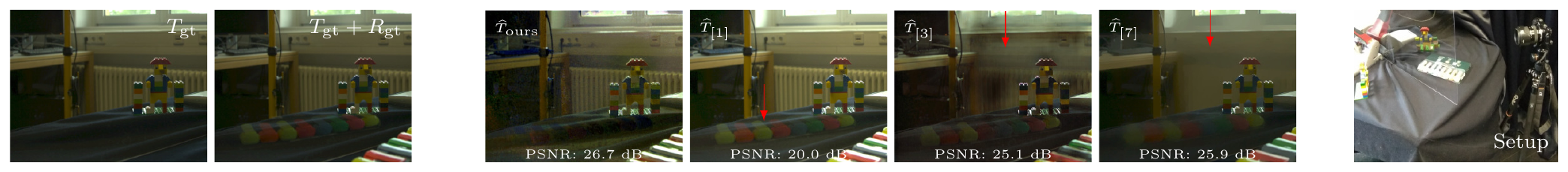}
\caption{By removing the semi-reflector, we can capture the ground truth transmission, $T_{\text{gt}}$, optically.}
\label{fig:numerical_real}
\end{figure}

Due to the need for ground-truth, a large-scale numerical evaluation can only be performed on synthetic data. For this task we take two datasets, the VOC2012~\cite{pascal-voc-2012} and the PLACE2~\cite{zhou2017places} datasets. A comparison with state-of-the-art methods shows that our method outperforms the second best method by a significant margin in terms of PSNR: $\sim 2$ dB, see Table~\ref{tab:synthetic_eval}.
For a numerical evaluation on real data, we set up a scene with a glass surface and objects causing reflections. After capturing polarization images of the scene, we removed the glass and captured the ground truth transmission, $T_{\text{gt}}$. Figure~\ref{fig:numerical_real} shows the transmission images estimated by different methods. Our method achieves the highest PSRN, and the least amount of artifacts.
\vspace{-2mm}

\subsection{Effect of Data Modeling}
\label{sec:data_modelling}

We also thoroughly validate our data-generation pipeline. Using both synthetic and real data, we show that the proposed non-rigid deformation (NRD) procedure and the local curvature generation (LCG) are effective and necessary. To do this, we train our network until convergence on three types of data: data generated only with the proposed dynamic range manipulation, DR for short, data generated with DR$+$NRD, and data generated with DR$+$NRD$+$LCG.

We evaluate these three models on a hold-out synthetic validation set that features all the transformations from Figure~\ref{fig:flow_chart_data}.
The table in Figure~\ref{fig:effect_bent} shows that the PSNR drops significantly when only part of our pipeline is used to train the network.
Unfortunately, a numerical evaluation is only possible when the ground truth is available.
However, Figure~\ref{fig:effect_bent} shows the output of the three models on the real image from Figure~\ref{fig:teaser}. The benefits of using the full pipeline are apparent.

A visual inspection of Figure~\ref{fig:teaser} allows to appreciate that, thanks to our ability to deal with curved surfaces and dynamic scenes, we achieve better performance than the state-of-the-art methods.

    \begin{minipage}[b]{0.5\linewidth}
    \centering
    \begin{tikzpicture}
    \node {\includegraphics[scale=.6]{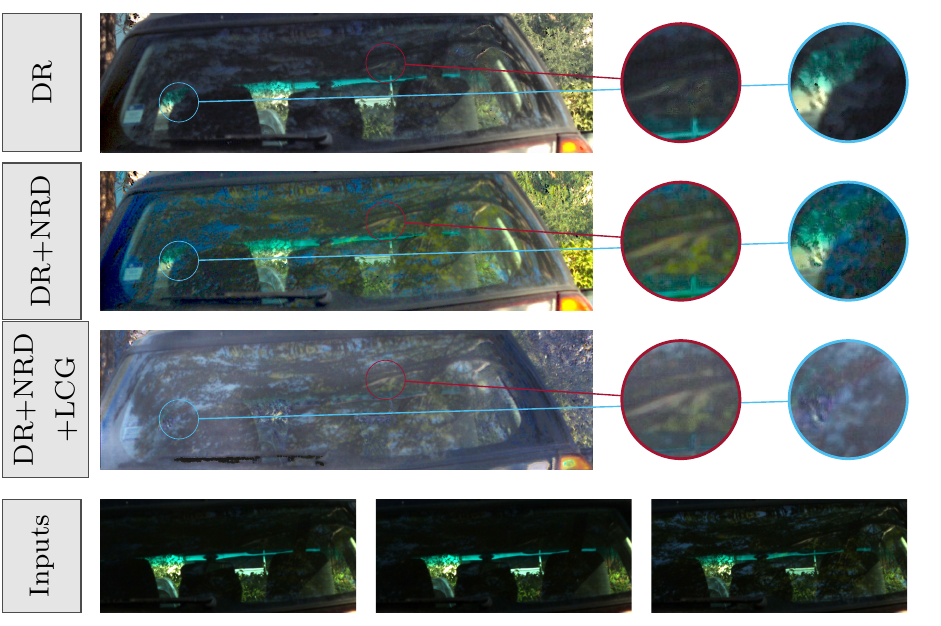}};

  \end{tikzpicture}
    \end{minipage}
    \hspace{1cm}
    \begin{minipage}[b]{0.3\linewidth}
    \centering

    \begin{tabular}{lccccc}
\toprule
Model & PSNR\\
\midrule
DR & 28.17 dB\\
DR+NRD & 30.44 dB\\
DR+NRD+LCG & 31.18 dB\\
\bottomrule
\end{tabular}
\label{tab:synth_evaluation}
\end{minipage}

\begin{minipage}[b]{0.95\linewidth}
\captionof{figure}{Our reflection estimation (left) on a real-world curved surface and synthetic data (right Table) using the same network architecture trained on different components of our data pipeline. Only when using the full pipeline (DR+NRD+LCG) the reflection layer is estimated correctly. Note how faint the reflection is in the inputs (bottom row).}
\label{fig:effect_bent}
\end{minipage}

\subsection{Evaluation on Real-World Examples}
\label{sec:evaluation_realworld}
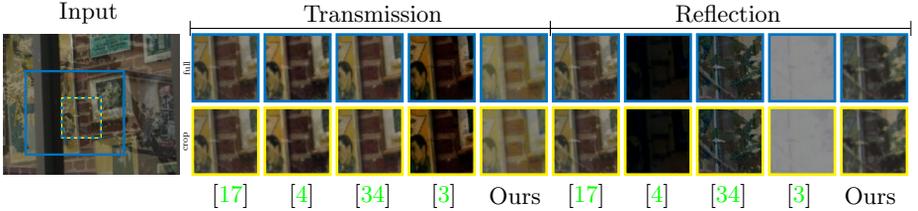
\begin{figure}[tb]
  \centering
  \begin{tikzpicture}
    \node{\includegraphics[width=\textwidth]{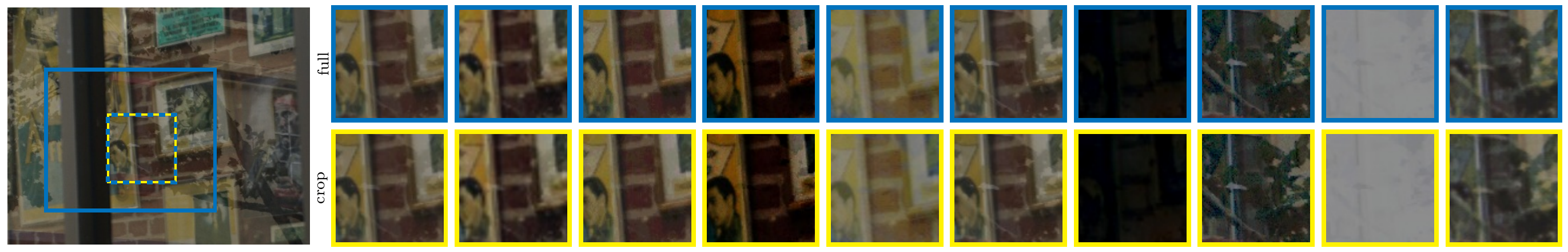}};
    \node at (-3.0 + 0*0.95, -1.2) {\cite{farid1999cvpr}};
    \node at (-3.0 + 1*0.95, -1.2) {\cite{fan2017iccv}};
    \node at (-3.0 + 2*0.95, -1.2) {\cite{schechner1999scia}};
    \node at (-3.0 + 3*0.95, -1.2) {\cite{kong2014pami}};
    \node at (-3.0 + 4*0.95, -1.2) {Ours};
    \node at (-3.0 + 5*0.95, -1.2) {\cite{farid1999cvpr}};
    \node at (-3.0 + 6*0.95, -1.2) {\cite{fan2017iccv}};
    \node at (-3.0 + 7*0.95, -1.2) {\cite{schechner1999scia}};
    \node at (-3.0 + 8*0.95, -1.2) {\cite{kong2014pami}};
    \node at (-3.0 + 9*0.95, -1.2) {Ours};

    \node at (-3.0 + -2*0.95, 1.2) {Input};
    \node at (-3.0 + 2*0.95, 1.2) {Transmission};
    \node at (-3.0 + 7*0.95, 1.2) {Reflection};

    \draw[{|-|}] (-3.55, 1) -- (1.28, 1) ;
    \draw[{-|}] (1.28, 1) -- (6.1, 1);

  \end{tikzpicture}
  \caption{Applying the different algorithms to the whole image and cropping a region (`full') is equivalent to applying the same algorithms to the cropped region directly (`crop').}
  \label{fig:pure_mixed}
\end{figure}
\begin{figure}[H]
  \centering
  \hspace*{-0.5cm}
  \resizebox{\linewidth}{!}{
  \begin{tikzpicture}
    \node {\includegraphics[width=12cm]{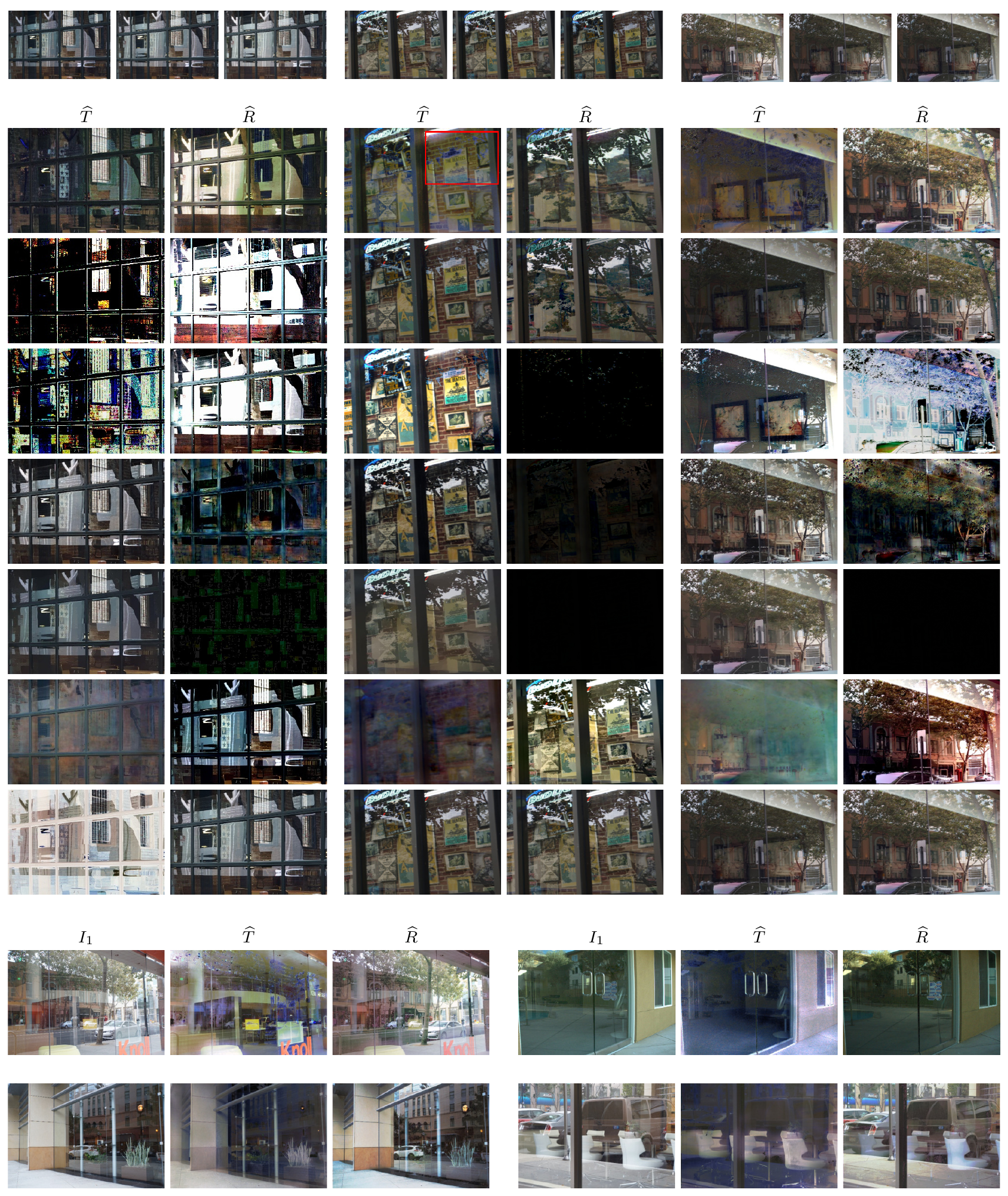}};

    \node at (-4, 7.5) {Curved Window};
    \node at (0, 7.5) {Bar};
    \node at (4, 7.5) {Paintings};

    \node[rotate=90] at (-6.5, 6.5) {Inputs};
    \node at (-6.5, 5) {Ours};
    \node at (-6.5, 3.68) {\cite{schechner2000polarization}};
    \node at (-6.5, 2.36) {\cite{kong2014pami}};
    \node at (-6.5, 1.04) {\cite{fan2017iccv}};
    \node at (-6.5, -0.28) {\cite{arvanitopoulos2017single}};
    \node at (-6.5, -1.60) {\cite{li2014single}};
    \node at (-6.5, -2.92) {\cite{farid1999cvpr}};

    \node[rotate=90] at (-6.2, -4.8) {Knoll};
    \node[rotate=90] at (0, -4.8) {Gym};
    \node[rotate=90] at (-6.2, -6.3) {Side Street};
    \node[rotate=90] at (0, -6.3) {Chairs};

  \end{tikzpicture}
  }
  \caption{Results on typical real-world scenes. Top pane: comparison with state-of-the-art methods, bottom pane: additional results. More results are given in the Supplementary.}
  \label{fig:large_comparison}
\end{figure}
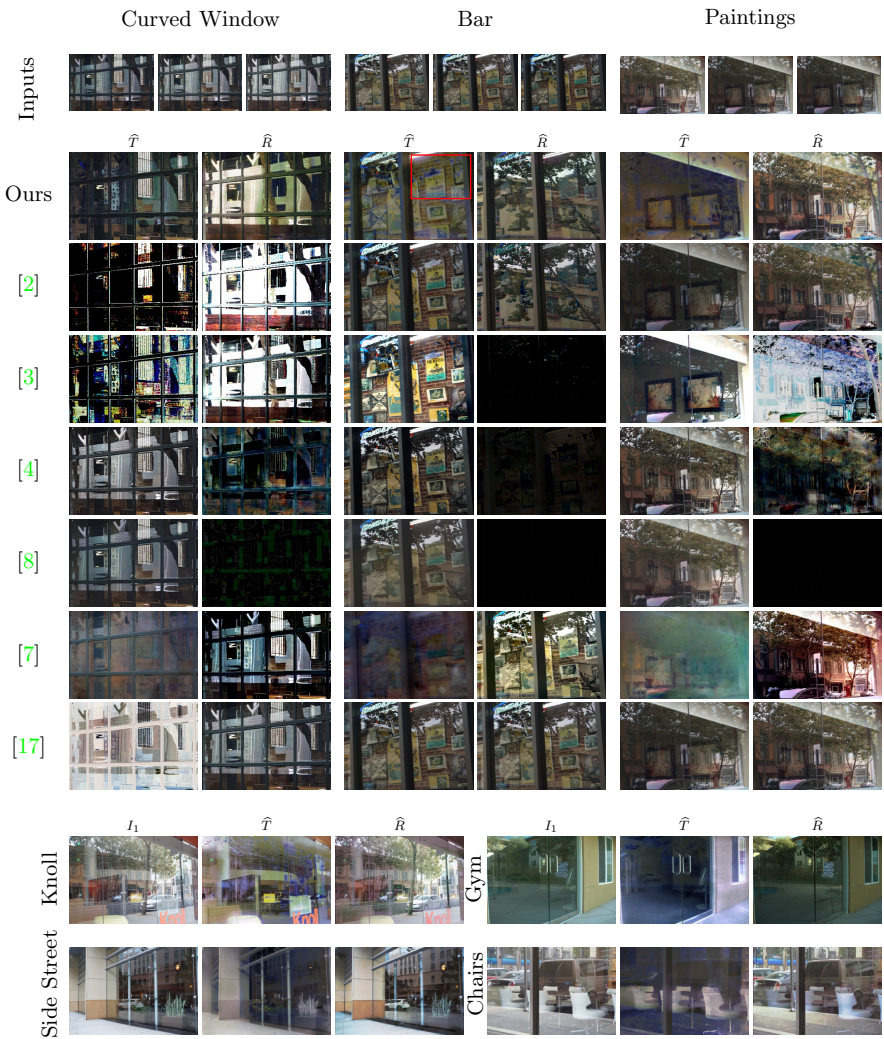

We extensively evaluate our method against previous work on the proposed URD.
For fairness towards competing methods, which make stronger assumptions or expect different input data, we slightly adapt them, or run them multiple times with different parameters retaining {}only the best result.
Due to space constraints, Figure~\ref{fig:large_comparison} only shows seven of the results.
We refer the reader to the Supplementary for the rest of the results and for a detailed explanation about how we adapted previous methods.
One important remark is in order. Although the images we use include opaque objects, \ie the semi-reflector does not cover the whole picture, the methods against which we compare are local: applying the different algorithms to the whole image and cropping a region is equivalent to applying the same algorithms to the cropped region directly, Figure~\ref{fig:pure_mixed}.

Figure~\ref{fig:large_comparison}, \textit{Curved Window} shows a challenging case in which the AOI is significantly different from $\theta_B$ across the whole image, thus limiting the effect of the polarizer in all of the inputs.
Moreover, the glass surface is slanted and locally curved, which breaks several of the assumptions of previous works. As a result, other methods completely fail at estimating the reflection layer, the transmission layer, or both.
On the contrary, our method separates \est{T} and \est{R} correctly, with only a slight halo of the reflection in \est{T}. In particular, notice the contrast of the white painting with the stars, as compared with other methods.
While challenging, this scene is far from uncommon.

Figure~\ref{fig:large_comparison}, \textit{Bar} shows another result on which our method performs significantly better than most related works. On this example, the method by Schechner~\etal~\cite{schechner2000polarization} produces results comparable to ours. However, recall that, to be fair towards their method, we exhaustively search the parameter space and hand-pick the best result.
Another thing to note is that our method may introduce artifacts in a region for which there is little or no information about the reflected or transmitted layer in any of the inputs, such as the case in the region marked with the red square on our \est{T}.

We also show an additional comparison showing the superiority of our method (Figure~\ref{fig:large_comparison}, \textit{Paintings}) and a few more challenging cases.
We note that in a few examples, our method may fail at removing part of the ``transmitted''  objects from \est{R}, as is the case in Figure~\ref{fig:large_comparison}, \textit{Chairs}.

\clearpage
\paragraph{User Study}
\begin{wraptable}{r}{0.41\textwidth}
  \caption{Result from the user study. We report the average recall-rate for each method.}
  \label{tab:recall}
  \centering

  \resizebox{\linewidth}{!}{
   \begin{tabular}{clllll}
  \toprule
                                   & \multicolumn{2}{c}{Transmission} &       & \multicolumn{2}{c}{Reflection}\\
  \cline{2-3}\cline{5-6}
  Method                           & $R@1$                            & $R@2$ &                                & $R@1$ & $R@2$\\
  \midrule
  Ours                             & \textbf{0.46}                            & \textbf{0.65} &                                  & \textbf{0.34} & \textbf{0.54}\\
  \cite{schechner2000polarization} & 0.14                            & 0.38 &                                  & 0.23 & 0.40\\
  \cite{kong2014pami}              & 0.11                            & 0.27 &                                  & 0.09 & 0.20\\
  \cite{fan2017iccv}               & 0.06                            & 0.17 &                                  & 0.08 & 0.20\\
  \cite{li2014single}              & 0.08                            & 0.21 &                                  & 0.10 & 0.29\\
  \cite{farid1999cvpr}             & 0.06                            & 0.13 &                                  & 0.15 & 0.37\\
  \bottomrule
  \end{tabular}
  }
\end{wraptable}
Since we do not have the ground truth for real data, we evaluate our method against previous results by means of a thorough user study. We asked $43$ individuals not involved with the project, to rank our results against the state-of-the-art~\cite{farid1999cvpr,fan2017iccv,li2014single,schechner2000polarization,kong2014pami}.
In our study, we evaluate \est{R} and \est{T} as two separate tasks, because different methods may perform better on one or the other.
For each task, the subjects were shown the three input polarization images, and the results of each method on the same screen, in randomized order. They were given the task to rank the results $1$--$6$, which took, on average, 35 minutes per subject.
We measure the recall rate in ranking, $R@k$, \ie the fraction of times a method ranks among the top-$k$ results. Table~\ref{tab:recall} reports the recall-rates.
Two conclusions emerge from analyzing the table. First, and perhaps expected, polarization-based methods outperform the other methods.
Second, our method ranks higher than related works by a significant margin.

\begin{comment}
\begin{figure}[tb]
  \centering
  \includegraphics[width=\textwidth]{figures/direct}
  \caption{Comparing the prediction of our network against an attempt directly estimating the different layers.
  Note, while a direct approach produces outputs which are reasonable in terms of objects, it cannot resolve the ambiguity between colors intensities from the reflection/transmission layer.
  Instead our network predicting additional blending mask provides correct colors as a consequence of this additional regularization.
  \PW{a sigmoid layer is very unlikely to produce either all zeros or all ones.}\TODO{PUT this at the correct place}}
  \label{fig:figure1}
\end{figure}
\end{comment}
\section{Conclusion}\label{sec:conclusion}

Separating the reflection and transmission layers from images captured \emph{in the wild} is still an open problem, as state-of-the-art methods fail on many real-world images.
Rather than learning to estimate the reflection and the transmission directly from the observations, we propose a deep learning solution that leverages the properties of polarized light: it uses a Canonical Projection Layer, and it learns the residuals of the reflection and transmission relative to the canonical images.
Another key ingredient to the success of our method is the definition of an image-synthesis pipeline that can accurately reproduce typical nonidealities observed in everyday pictures.
We also note that the non-rigid deformation procedure that we propose can be used for other stack-based methods where non-static scenes may be an issue.
To evaluate our method, we also propose the Urban Reflection Dataset, which we will make available upon publication.
Using this dataset, we extensively compare our method against a number of related works, both visually and by means of a user study, which confirms that our approach is superior to the state-of-the-art methods.
Finally, the code for most of the existing methods that separate reflection and transmission is not available: to perform an accurate comparison, we re-implemented representative, state-of-the-art works, and make our implementation of those algorithms available to the community, to enable more comparisons.

\section*{Acknowledgments}
We thank the reviewers for their feedback, in particular the reviewer who suggested the experiment in Fig.~\ref{fig:numerical_real}, Hendrik P.A. Lensch for the fruitful discussions, and the people who donated half hour of their lives to take our survey.
\clearpage
\bibliographystyle{splncs}
\bibliography{egbib}
\end{document}